\documentclass[runningheads]{llncs}
\usepackage{graphicx}

\usepackage{tikz}
\usepackage{comment}
\usepackage{amsmath,amssymb} 
\usepackage{color}
\usepackage[mathscr]{euscript}

\usepackage{multirow}
\usepackage{multicol}
\usepackage{makecell}

\usepackage{wrapfig}

\usepackage[pagebackref,breaklinks,colorlinks]{hyperref}
\usepackage{booktabs} 

\usepackage[capitalize]{cleveref}
\crefname{section}{Sec.}{Secs.}
\Crefname{section}{Section}{Sections}
\Crefname{table}{Table}{Tables}

\usepackage[accsupp]{axessibility}  

\newcommand{\et}[2]{${#1}^{\pm{#2}}$}
\newcommand{\etb}[2]{$\mathbf{{#1}}^{\pm{#2}}$}
\newcommand{\ets}[2]{$\underline{{#1}}^{\pm{#2}}$}

\newcommand{\beforefigcaption}{\vspace{-9mm}}
\newcommand{\afterfigcaption}{\vspace{-5mm}}

\newcommand{\aftertab}{\vspace{-9mm}}
\newcommand{\beforesection}{\vspace{-2mm}} 
\newcommand{\aftersection}{\vspace{-2mm}}
\newcommand{\beforesubsection}{\vspace{-2mm}}
\newcommand{\aftersubsection}{\vspace{-2mm}}
\newcommand{\beforesubsubsection}{\vspace{-5mm}}

\begin{document}
\pagestyle{headings}
\mainmatter
\def\ECCVSubNumber{650}  

\title{TM2T: Stochastic and Tokenized Modeling for the Reciprocal Generation of 3D Human Motions and Texts} 

\titlerunning{TM2T: Reciprocal Generation of 3D Human
Motions and Texts}
%
\author{Chuan Guo\orcidID{0000-0002-4539-0634} \and Xinxin Zuo \and Sen Wang \and Li Cheng}
\authorrunning{Chuan Guo et al.}
%
\institute{University of Alberta \\
\email{\{cguo2, xzuo, sen9, lcheng5\}@ualberta.ca}}

\maketitle

\begin{abstract}
Inspired by the strong ties between vision and language, the two intimate human sensing and communication modalities, our paper aims to explore the generation of 3D human full-body motions from texts, as well as its reciprocal task, shorthanded for text2motion and motion2text, respectively. To tackle the existing challenges, especially to enable the generation of multiple distinct motions from the same text, and to avoid the undesirable production of trivial motionless pose sequences, we propose the use of motion token, a discrete and compact motion representation. This provides one level playing ground when considering both motions and text signals, as the motion and text tokens, respectively. 
Moreover, our motion2text module is integrated into the inverse alignment process of our text2motion training pipeline, where a significant deviation of synthesized text from the input text would be penalized by a large training loss; empirically this is shown to effectively improve performance. Finally, the mappings in-between the two modalities of motions and texts are facilitated by adapting the neural model for machine translation (NMT) to our context. This autoregressive modeling of the distribution over discrete motion tokens further enables non-deterministic production of pose sequences, of variable lengths, from an input text. 
Our approach is flexible, could be used for both text2motion and motion2text tasks. 
Empirical evaluations on two benchmark datasets demonstrate the superior performance of our approach on both tasks over a variety of state-of-the-art methods. Project page: \href{https://ericguo5513.github.io/TM2T/}{https://ericguo5513.github.io/TM2T/}


\keywords{Motion captioning, text-to-motion generation}
\end{abstract}

\begin{figure*}[t]
	\centering
	\includegraphics[width=\linewidth]{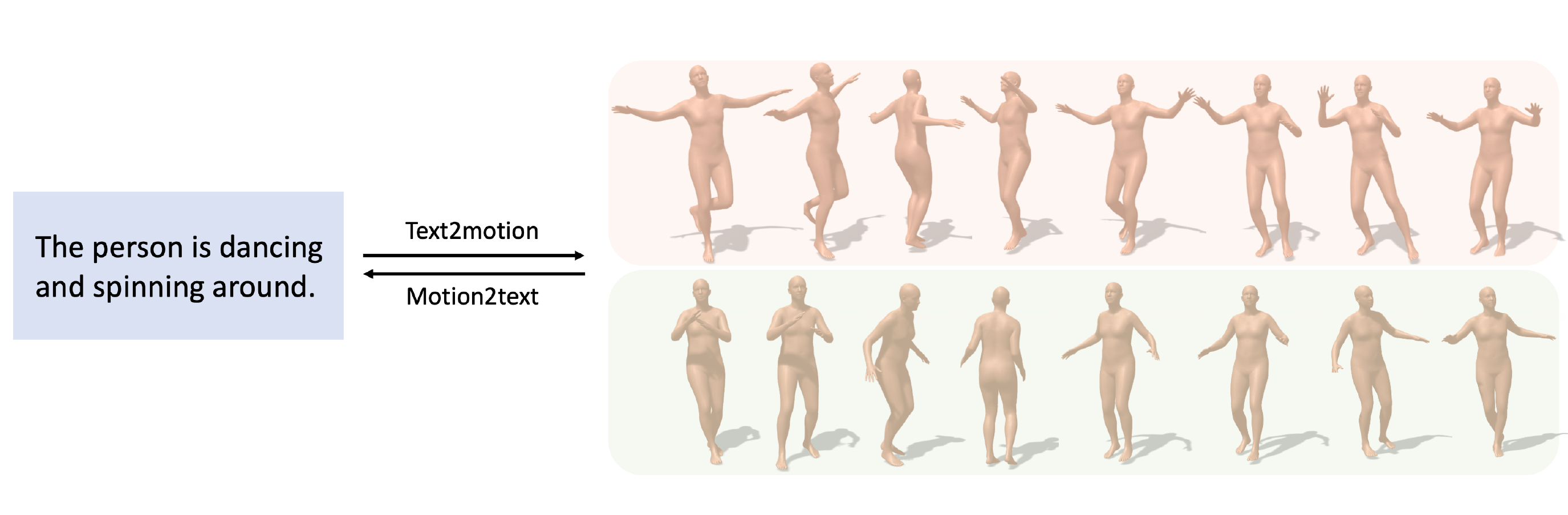}
		 \beforefigcaption
	\caption{An illustration of our bidirectional TM2T approach that captures the interplay between text (left) and 3D human motion (right) through the text2motion and motion2text modules. Note the stochastic nature of our text2motion module allows the generation of different 3D motions from the same textural description. }
	\label{fig:teaser_image}
	 \afterfigcaption
\end{figure*}

\beforesection
\section{Introduction}
\aftersection

The interplay of vision and language is important in our daily life and social functions. It has motivated considerable research progresses in related topics such as image or video captioning~\cite{xu2015show,vinyals2015show}, and language grounded generation of images or videos~\cite{zhang2017stackgan,xu2018attngan,li2018video}. On the other hand, when coming to human motion analysis, the connections between visual and textural aspects of human motions are much less studied. 
Existing efforts primarily focus on unidirectional mapping of either motion captioning (motion2text)~\cite{goutsu2021linguistic,takano2015statistical} or language grounded motion generation (text2motion)~\cite{ahuja2019language2pose,ghosh2021synthesis,lin2018generating}, with only two~\cite{plappert2018learning,yamada2018paired} exploring the integration of visual 3D motions and their textural descriptions. 
However, both studies tend to produce static pose sequences when motion lengths are longer than 3-4 seconds. Both requires as input the initial pose \& target motion length. They are also deterministic methods. That is, each of them always generates the same motions from a given text script. 
The first phenomenon of lifeless motions could be largely attributed to the direct use of raw 3D poses as their motion representation, which is unnecessarily redundant and yet fails to capture the local contexts of the underlying motion dynamics. The second issue is rooted in their deterministic motion generation processes, that are in contrary to our daily experiences, where multiple distinct motion styles often exist for a character to perform under a same textural script. The conditioning on initial state and target length further imposes strict constraint toward being practically feasible.



The aim of this paper is to investigate the bi-directional ties between 3D human full-body motions and their language descriptions, as illustrated in~\cref{fig:teaser_image}. 
Given the asymmetric nature of the two underlying tasks, where text2motion is typically a much harder problem than the reciprocal task, motion2text, our primary focus is text2motion, with a secondary emphasis on motion2text. It is worth noting that in our approach, the module (also called motion2text for simplicity) developed for motion2text task, is also utilized as an integral part of our text2motion training process, referred to as inverse alignment in Fig.~\ref{fig:framework}(c). Empirical evidences suggest the benefit of this strategy in improving our performance for the text2motion problem.
To address the lifeless motion issue, we introduce motion token, a compact and semantically rich representation for 3D motions. This is achieved by adapting the deep vector quantization~\cite{van2017neural} in our context to learn a spatial-temporal codebook from the 3D pose sequences in the training set, with each entry in the codebook describing a particular kind of motion segments. 3D motions are then reconstructed by decoding the compositions of a list of codebook entries. This way, a 3D human motion is represented as a list of motion tokens (i.e. discrete indices to the codebook entries), each encoding its local spatial-temporal context. This discrete representation also facilitates the follow-up neural machine translators (NMTs)~\cite{vaswani2017attention,Bahdanau2014Neural} to construct mappings between the stream of motion tokens from the motion side, and the stream of text tokens from the language side. Furthermore, our proposed approach is able to explicitly model the underlying distribution of 3D motions conditioned on texts, instead of regressing the mean motions as in previous works ~\cite{ahuja2019language2pose,ghosh2021synthesis,lin2018generating,yamada2018paired,plappert2018learning}, thus allows non-deterministic text2motion generation.


Our main contributions can be summarized as follows: (i) a motion token representation that compactly encodes 3D human motions. Together with the other key ingredients, including NMT mappings in-between the motion-token and text-token sequences, the motion2text-based inverse alignment, as well as the distribution sampling for non-deterministic predictions, our approach is capable of generating 3D motions (i.e. pose sequences) that are distinct in their lengths and styles, visually pleasing, and importantly, semantically faithful to the same input script. 
Our approach is also flexible, in that it can be use for both text2motion and motion2text tasks. (ii) Extensive empirical evaluations over two motion-language benchmark datasets demonstrate the superior performance of our approach over a variety of state-of-the-art methods when examined on each of the two tasks. 

\beforesection
\section{Related Work}
\aftersection

\noindent\textbf{Image/Video Captioning and Motion2text.}
Vision grounded text generation has a long history with extended literature. Here we only focus on the closely related topic of image and video captioning. Early methods~\cite{kulkarni2013babytalk,kojima2002natural} commonly approach this problem by tagging parts of sentences such as nouns and verbs from visual contents, followed by filling in pre-defined sentence templates. With the advent of deep neural networks, the tools used for visual captioning have been significantly changed. Take~\cite{vinyals2015show} for example, it starts by extracting high-level image features from pre-trained GoogleNet, which are then fed into a LSTM decoder to produce captions. 
In \cite{venugopalan2015sequence}, an RNN-based video captioning model is considered, that extracts individual frame features from pre-trained CNN, and translates them to sentences through sequence-to-sequence learning. Further extensions are made through e.g. incorporating attention mechanism for better vision-language alignment~\cite{xu2015show,wang2016temporal}. More recent methods consider the use of various deep learning apparatus such as GANs~\cite{guo2019mscap,park2019adversarial}, deep reinforcement learning~\cite{gao2019self,qin2019look}, and transformers~\cite{dubey2021label,ging2020coot}. 

In contrast, research efforts on captioning 3D human motions are considerably more limited. \cite{takano2015statistical} learns the mapping from human motions to language relying on two statistical models: one associates motions with words; the other assembles words back to form sentences. Recurrent networks are utilized by \cite{yamada2018paired,plappert2018learning} to address this task. In \cite{yamada2018paired}, motion and text features are extracted by two autoencoders respectively; this is followed by generating texts and motions from each other through shared latent vectors. Sequence-to-sequence RNNs are adopted in~\cite{plappert2018learning} to translate motions to scripts. Recently, the work of \cite{goutsu2021linguistic} proposes SeqGAN that extends NMT model with a discriminator. Some common issues with existing motion2text results are typically short in length, often incomplete in content, and sometimes lack in details. 

\noindent\textbf{Human Motion Modeling and Text2motion.}
The importance of human motion modeling has been manifested through the extensive research efforts in recent years, where motions are produced based on various forms of inputs, such as partial pose sequences, 
control signals, action category, and text. Future motion prediction aims to generate short \cite{xu2017lie} and long~\cite{liu2019towards,pavllo2020modeling} future pose sequences based on partial pose sequences. This has been traditionally modeled in one-to-one mapping fashion until recent works~\cite{aliakbarian2021contextually,yuan2020dlow,mao2021generating} that take account the stochastic nature of human motion dynamics. The efforts of~\cite{wang2021multi,adeli2020socially,cao2020long,corona2020context} proceed to predict multi-person or scene-aware 3D motions. Meanwhile, \cite{holden2017phase,holden2020learned,starke2019neural} attempts to model human motions according to instant control signals such as velocity and directional readouts. 
In \cite{holden2017phase}, feet contact information is fed into a phase function to produce blending weights of four expert MLP networks. The blended MLP network then predicts next pose state given current state and goal control signals. This is extended in \cite{van2017neural,holden2020learned} where the phase function is replaced by a learnable gating network. Action category based human motion generation also draws considerable interests by resorting to a diverse range of learning strategies, including GANs~\cite{wang2020learning}, VAEs~\cite{guo2020action2motion,guo2022action2video}, Transformers~\cite{petrovich2021action} and GCNs~\cite{yu2020structure}. 

In terms of text based human motion modeling (text2motion), the sequence-to-sequence RNN models have been considered by~\cite{lin2018generating,plappert2018learning}; in \cite{ahuja2019language2pose}, a latent embedding space is proposed, which is shared by both text and pose sequences and is trained via curriculum learning. The work of \cite{ghosh2021synthesis} considers the topology of human skeleton, and proposes a hierarchical two-stream pose generator. Note existing techniques developed in text2motion are predominantly deterministic. 
This is in contrast to our proposed stochastic motion generation process. 

\noindent\textbf{Discrete Vector Quantization.} \cite{van2017neural} advocates the quantization of continuous features into discrete latent representation by training a variational autoencoder. This is followed up by several more recent efforts to improve the representation quality and reconstruction accuracy, including hierarchical feature representation~\cite{razavi2019generating}, gumbel-softmax relaxation~\cite{ramesh2021zero} and adversarial training~\cite{esser2021taming}. In \cite{peng2021generating}, hierarchical vector quantization is carried out in encoding and generating diverse image patches for inpainting; the work of \cite{rakhimov2020latent} leverages quantized video frame representation to synthesize future frames. These prior arts inspire the motion token scheme considered in our approach.

\beforesection
\section{Our Approach}
\aftersection

In what follows, we first detail how discrete motion tokens are obtained from raw 3D motions via vector quantization in \cref{app:m_vq_vae}. Based on this new motion representation, autoregressive NMT networks are used for modeling the bi-modal mappings of motion2text (\cref{app: motion-to-text}) and text2motion (\cref{app:text-to-motion}), with inverse alignment elaborated in \cref{app:text-to-motion}.

\beforesubsection
\subsection{Motion Tokens}
\aftersubsection
\label{app:m_vq_vae}

\begin{figure*}[t]
	\centering
	\includegraphics[width=\linewidth]{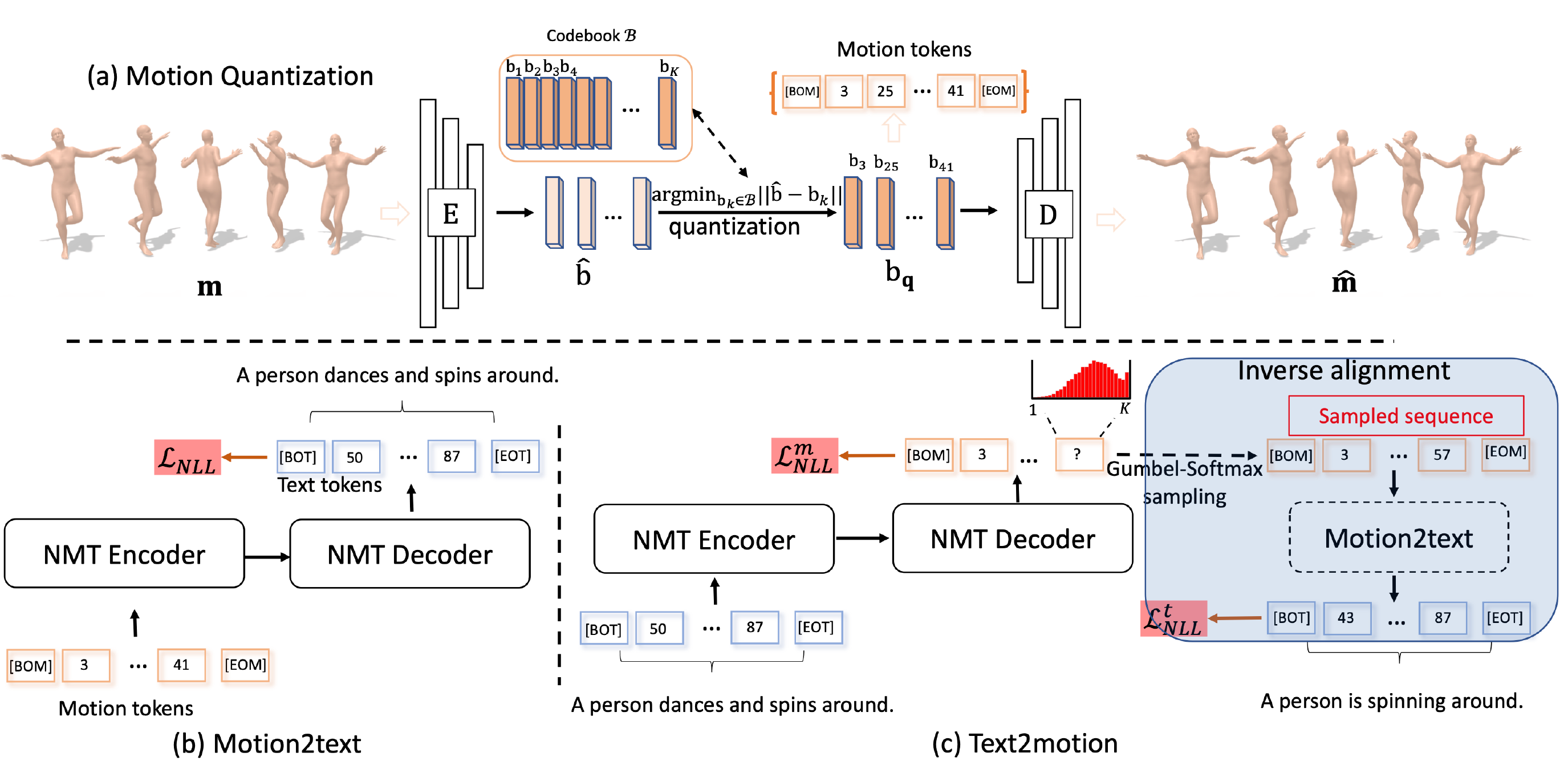}
		 \beforefigcaption
	\caption{\textbf{Approach overview.} (a) A 1D CNN based latent quantization model is firstly learned to reconstruct training motions. After training, a motion can be subsequently converted to a tuple of discrete motion tokens (i.e., codebook-indices). [BOM] and [EOM] are indicators of start and end added in a motion token sequence. (b-c) Mappings between motion and text tokens are modeled by autoregressive NMT networks and optimized by maximizing the log-likelihood of the targets ($\mathscr{L}_{NLL}$ and $\mathscr{L}_{NLL}^m$). (c) While training text2motion, motion tokens sampled from the resulting discrete distributions are inversely mapped to the text space via the learned motion2text model. Loss $\mathscr{L}_{NLL}^t$ penalizes the inverse alignment error. Finally, the 3D pose sequence is obtained by decoding motion tokens via the decoder $\mathrm{D}$ in (a).}
	\label{fig:framework}
	 \afterfigcaption
\end{figure*}

We pre-train a latent quantization model on 3D human motions as presented in Fig.~\ref{fig:framework} (a). Given the pose sequence $\mathbf{m} \in \mathbb{R}^{T\times D_p}$, where $T$ denotes the number of poses and $D_p$ pose dimension, a series of 1D convolutions are applied along the time (i.e. 1st) dimension that yields latent vectors $\mathbf{\hat{b}}\in \mathbb{R}^{t\times d} (t<T)$ with $d$ being number of convolution kernels. This process could be written as $\mathbf{\hat{b}}=\mathrm{E}(\mathbf{m})$. 

Then, $\mathbf{\hat{b}}$ is transformed to a collection of codebook entries $\mathbf{b}_{\mathbf{q}}\in \mathbb{R}^{t\times d}$ through discrete quantization. Specifically, the learnable codebook $\mathscr{B}=\{\mathbf{b}\}_{k=1}^K\subset \mathbb{R}^d$ consists of $K$ latent embedding vectors with each a $d$-dimensional vector. The process of quantization $\mathrm{Q}(\cdot)$ is operated by replacing each 
row vector $\mathbf{\hat{b}}_i\in \mathbb{R}^d$ in $\mathbf{\hat{b}}$ with its nearest codebook entry $\mathbf{b}_k$ in $\mathscr{B}$, defined as
\begin{align}
\label{eq:quantization}
    \mathbf{b}_{\mathbf{q}} = \mathrm{Q}(\mathbf{\hat{b}}):=\left(\mathrm{argmin}_{\mathbf{b}_k \in \mathscr{B}} \|\mathbf{\hat{b}}_i - \mathbf{b}_k\| \right) \in \mathbb{R}^{t\times d}.
\end{align}

\begin{figure*}[t]
	\centering
	\includegraphics[width=\linewidth]{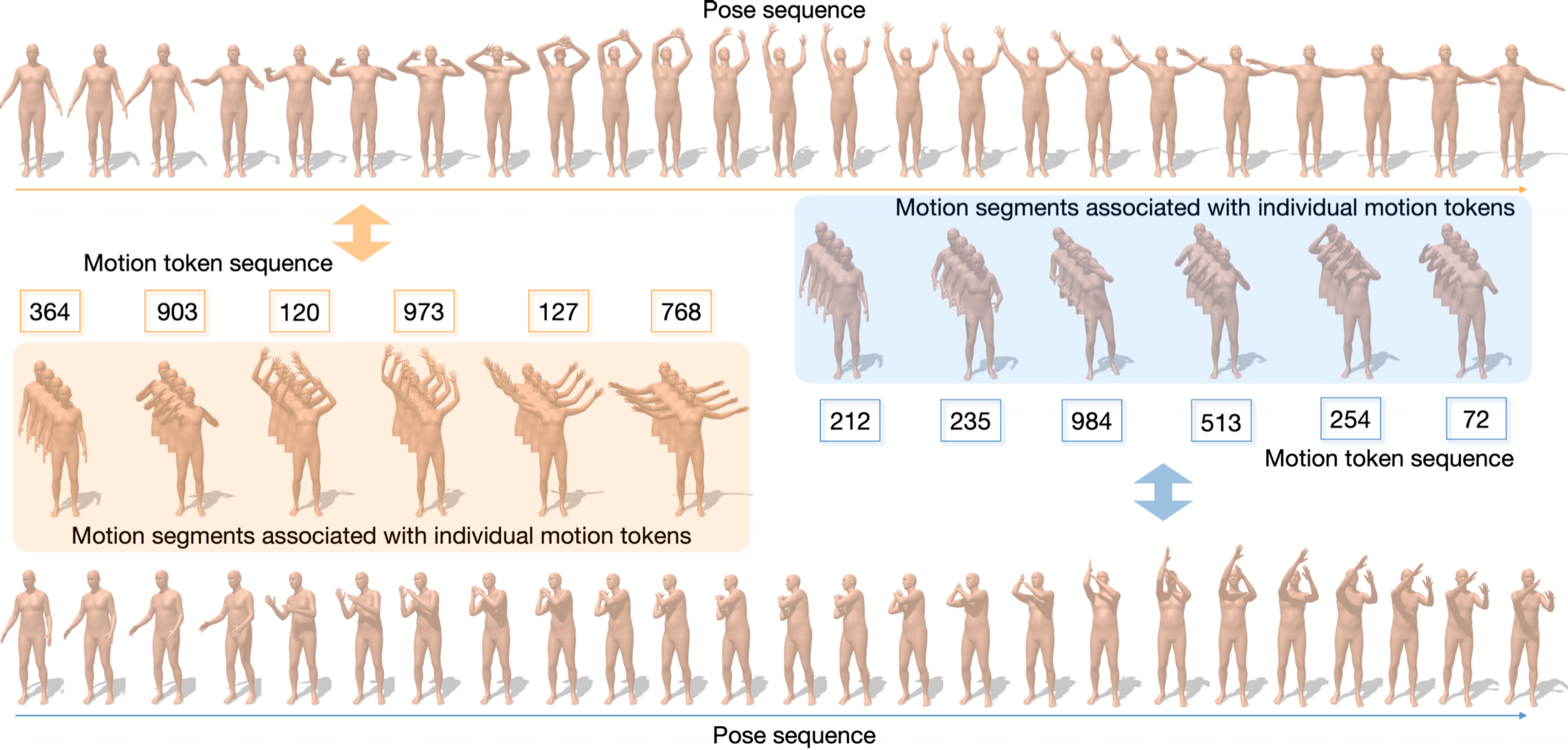}
		 \beforefigcaption
	\caption{Exemplar results of motion tokens (middle) and their corresponding pose sequences (top and bottom). Here two 24-frame pose sequence examples are presented; each is reconstructed from a motion token sequences of size 6. Each motion token is associated with a specific local spatial-temporal context, visualized in 4-frame motions.}
	\label{fig:token_visualization}
	 \afterfigcaption
\end{figure*}

A following de-convolutional decoder $\mathrm{D}$ projects $\mathbf{b}_{\mathbf{q}}$ back to the 3D motion space as a pose sequence, $\hat{\mathbf{m}}$. Now, the entire process can be formulated as
\begin{align}
    \hat{\mathbf{m}} = \mathrm{D}(\mathbf{b}_\mathbf{q}) = \mathrm{D}(\mathrm{Q}(\mathrm{E}(\mathbf{m}))).
\end{align}
This is trained via a reconstruction loss combined with embedding commitment loss terms that encourage latent alignment and stabilize training process:
\begin{align}
    \mathcal{L}_{vq} = \|\mathbf{\hat{m}} - \mathbf{m}\|_1 + \|\mathrm{sg}[\mathrm{E}(\mathbf{m})] - \mathbf{b}_\mathbf{q}\|_2^2 + \beta\|\mathrm{E}(\mathbf{m}) - \mathrm{sg}[\mathbf{b}_\mathbf{q}]\|_2^2,
\end{align}
where $\mathrm{sg}[\cdot]$ denotes the stop-gradient operation, and $\beta$ a weighting factor. Straight-through gradient estimator~\cite{van2017neural} is employed to allow gradient backpropagation through the non-differentiable quantization operation in Eq.\eqref{eq:quantization} that simply copies the gradients from the decoder $\mathrm{D}$ to the encoder $\mathrm{E}$. 

During inference, a pose sequence $\mathbf{m}\in \mathbb{R}^{T\times D_p}$ can be represented as a sequence of discrete codebook-indices $s\in\{1,...,|\mathscr{B}|\}^t$ (namely \textit{motion tokens}) of quantized embedding vectors $\mathbf{b}_{\mathbf{q}}$, where $s_i=k$ such that $(\mathbf{b}_\mathbf{q})_i = \mathbf{b}_k$. By mapping motion tokens back to their corresponding codebook entries $\mathbf{b}_\mathbf{q} = (\mathbf{b}_{s_i})$, human poses are then readily recovered using decoder $\mathbf{\hat{m}} = \mathrm{D}(\mathbf{b}_\mathbf{q})$. [BOM] and [EOM] are respectively added to the start and end of a motion token sequence as boundary indicators.

\beforesubsubsection
\subsubsection{Motion Token Contexts.} With vector quantization, each motion token is associated with a particular type of motion contexts, thus a 3D motion can be regarded as a meaningful composition of motion tokens. We decode each entry in the learned codebook $\mathscr{B}$ using decoder $\mathrm{D}$ and get 4-frame motion segments ($t=\frac{T}{4}$ in our setting) that reflect the contexts associated with individual motion tokens. Fig.~\ref{fig:token_visualization} presents two raw pose sequences and their motion token representations, as well as the associated motion segments. We can observe that, with global dependencies maintained in motion token sequences, each motion token successfully captures the spatial-temporal characteristics in local contexts. 

\beforesubsection
\subsection{Learning Motion2text}
\aftersubsection
\label{app: motion-to-text}
Given tokenized motion representation, we are able to efficiently build mapping from human motions to texts using NMT models such as Transformer~\cite{vaswani2017attention}. Assume the target is a sequence of text tokens $x\in\{1,...,|\mathscr{V}|\}^N$, where $\mathscr{V}$ is the word vocabulary and $N$ number of words in the description. As described in Fig.~\ref{fig:framework} (b), source motion tokens are fed into Transformer encoder and then the decoder predicts the probability distribution of possible discrete text tokens at each step $p_\theta(x|s)=\prod_i p_\theta(x_i|x_{<i}, s)$. Thus the training goal is to maximize the log-likelihood of the target sequence,
\begin{align}
    \mathcal{L}_{NLL} = -\sum_{i=0}^{N-1} \log p_\theta(x_i|x_{<i}, s).
\end{align}

\beforesubsection
\subsection{Learning Text2motion}
\aftersubsection
\label{app:text-to-motion}

Similarly, generating motions from language description can be modeled as autoregressive next-token predictions conditioned on textual inputs. Here we investigate two NMT models as our backbone: attentive GRU and Transformer, and examine our idea of \textit{inverse alignment} on GRU-based model. Since Transformer is typically trained with full teacher force, optimizing the Transformer-based text2motion with inverse alignment is extremely complicated. In other words, every time when generating the density function of next motion token, we need to input the whole history to the Transformer decoder and feed forward. As a result, to sample a complete motion token sequence, the computational (or optimization) graph will be extremely high. Therefore, we specifically introduce the procedure of using GRU based model as an example.

As is shown in Fig.~\ref{fig:framework} (c), firstly, a bi-directional GRU (i.e., NMT Encoder) models the temporal dependencies in language $x\in\{1,...,|\mathscr{V}|\}^N$, and produces sentence feature vector $\mathbf{s}\in \mathbb{R}^{d_l}$ as well as word feature vectors $\mathbf{w}\in \mathbb{R}^{N\times d_l}$, with $d_l$ denoting the dimensionality of hidden vectors. The NMT decoder, modeled as attention-based GRU, processes $\mathbf{s}$ and $\mathbf{w}$ and predicts the probability distribution over discrete motion tokens $\{1,...,|\mathscr{B}|\}$ autoregressively. In particular, GRU decoder is initialized by sentence vector $\mathbf{s}$, and then takes the attention vector $\mathbf{w}_{att}$ together with motion token as input at each time step. The attention vector $\mathbf{w}_{att}^t$ at time $t$ is obtained via
\begin{align}
    \mathbf{Q} &= \mathbf{h}_{t-1}\mathbf{W}^Q, \mathbf{K}=\mathbf{w}\mathbf{W}^K, \mathbf{V}=\mathbf{w}\mathbf{W}^V,\\
    \mathbf{w}_{att}^t&=\mathrm{softmax}\left(\frac{\mathbf{QK}^T}{\sqrt{d_{att}}}\right)\mathbf{V},
\end{align}
where $\mathbf{h}_{t-1}\in \mathbb{R}^{d_h}$ is previous hidden state in decoder, $\mathbf{W}^K, \mathbf{W}^V\in \mathbb{R}^{d_l\times d_{att}}$ and $\mathbf{W}^Q\in \mathbb{R}^{d_h\times d_{att}}$ are trainable weights with $d_h$ and $d_{att}$ denoting the dimension of hidden unit and attention vector respectively. During generation, motion tokens are sampled from predicted distribution $p_\phi(s_i|s_{<i},x)$ recursively until the end token (i.e., [EOM]) comes with maximum probability.

\beforesubsubsection
\subsubsection{Inverse Alignment.} Here we re-utilize the motion2text model in Sec.~\ref{app: motion-to-text} to further align the semantics between texts and generated motions. In detail, motion token sequence $\hat{s}$ is sampled from the approximated distribution $p_\phi(s|x)$, which is taken as input to the learned motion2text model and mapped to language tokens $x$ with probability $p_\theta(x|\hat{s})$. Note motion2text model is no longer updated here. However, sampling from discrete distribution is non-differentiable that does not allow the gradients back-propagating to the text2motion encoder and decoder. We instead resort to \textit{Gumbel-Softmax} reparameterization trick~\cite{jang2016categorical} to approximate the discrete sampling process. As the temperature $\tau$ of Gumbel-Softmax approaches 0, the resulting Gumbel-Softmax distribution becomes identical to the discrete distribution $p_\phi(s_i|s_{<i},x)$ and the sampled vectors become one-hot.

In summary, the final training objective turns to be
\begin{align}
    \mathcal{L} = -\left(\sum_{i=0}^{K-1}\log p_\phi(s_i|s_{<i}, x) + \sum_{i=0}^{N-1}\log p_\theta(x_i|x_{<i}, \hat{s})\right).
\end{align}

3D pose sequences can finally be obtained by decoding sampled motion tokens $\hat{s}$ using quantization decoder $\mathrm{D}$ as described in Sec.~\ref{app:m_vq_vae}. With discrete motion tokens and autoregressive modeling, variable motion lengths are implicitly modeled by text2motion, that the NMT model particularly learns to predict the end token i.e. [EOM] with maximum probability as signal of termination. Moreover, our proposed approach is easy to train, and does not suffer from the known shortcomings in GAN and VAE such as "mode collapse".

\beforesection
\section{Experiments}
\aftersection

Extensive experiments are conducted to evaluate our learned motion2text (Sec.~\ref{exp:motion-to-text}) and text2motion mapping models(Sec.~\ref{exp:text-to-motion}).

\beforesubsection
\subsection{Datasets}
\aftersubsection
Two 3D human motion-language datasets are considered for evaluation:
\begin{itemize}
    \item \textit{HumanML3D}~\cite{guo2022generating} is a large 3D human motion dataset that covers a broad range of human actions such as locomotion, sports, and dancing. It consists of 14,616 motions and 44,970 text descriptions. Each motion clip comes with at least 3 descriptions. Motions are re-scaled to 20 frames per second (FPS), resulting in duration ranges from 2 to 10 seconds. 
    \item \textit{KIT Motion-Language}~\cite{plappert2016kit} contains 3,911 3D human motion clips and 6,278 text descriptions. For each motion, the corresponding number of text descriptions ranges from one to four. Following~\cite{ahuja2019language2pose,ghosh2021synthesis}, these pose sequences are all sub-sampled to 12.5 FPS.
\end{itemize}

Both datasets are split into training, testing and validation sets with ratio of 0.8:0.15:0.05, which are further augmented by mirroring motions and replacing corresponding words in their text descriptions (e.g., 'left'$\rightarrow$'right').

\beforesubsection
\subsection{Metrics}
\aftersubsection

Besides traditional measurements, we also manage to evaluate the correspondences between motion and language using deep multimodal features. In particular, we train a simple framework that engages a motion feature extractor and a text feature extractor under contrastive assumption, that learn to produce geometrically closed feature vectors for matched text-motion pairs, and vice versa. Further details are relegated to supplementary file due to limited space. 


\textbf{R-Precision and Multimodal Distance} are proposed to gauge how well a text and a motion are semantically aligned. Take the evaluation of motion2text mapping for an example. For each generated description, we take its corresponding motion as well as 31 randomly selected mismatched motions from the test set as a motion pool. With text and motion feature extractors available, Euclidean distances between the description feature and each motion feature in the pool are calculated and ranked. The ground truth entry falling into the top-k (k=1,2,3) candidates is regarded as a successful retrieval. Then we count the average accuracy at top-k places, known as \textit{top-k R-precision}. Meanwhile, \textit{multimodal distance} is computed as the average Euclidean distance between text feature of each generated description and motion feature of its corresponding motion in the test set. Computing R-precision and multimodal distance for text2motion mapping is analogically carried out except generated motions and ground truth description are accordingly used. 

Overall, an extensive set of metrics including Bleu~\cite{papineni2002bleu}, Rouge~\cite{lin2004rouge}, Cider~\cite{vedantam2015cider}, BertScore~\cite{zhang2019bertscore}, R Precision and multimodal distance are adopted to quantitatively measure the performance of our motion2text mapping. For evaluation of non-deterministic text2motion mapping, we primarily follow \cite{guo2022action2video} which uses Frechet Inception Distance (FID), diversity and multimodality, and our complementary metrics, R precision and multimodal distance. Details of metrics are deferred to be presented in supplementary file.

\beforesubsection
\subsection{Evaluation of Motion-to-text Translation}
\aftersubsection
\label{exp:motion-to-text}
We adopt RAEs~\cite{yamada2018paired} and SeqGAN~\cite{goutsu2021linguistic} as our baseline methods; RAEs~\cite{yamada2018paired} learns a shared embedding space for language and human motions via two recurrent autoencoders, while SeqGAN~\cite{goutsu2021linguistic} combines recurrent sequence-to-sequence model with a discriminator that judges whether a sentence is real or not. We further equip the vanilla RNN model in Seq2Seq~\cite{plappert2018learning} with late attention as another strong baseline (termed as Seq2Seq(Att)). A variant of our method not using motion tokens (ours w/o MT) is also engaged to analyze the role of motion token. Note that grammatical tense and plural of words are neglected in our setting in order to ease the learning process. Descriptions are produced using beam search strategy with size of 2 throughout all experiments.

\begin{table*}[t]
    \centering
    \scalebox{0.86}{

    \begin{tabular}{l l c c c c c c c c c}
    \toprule
    \multirow{2}{*}{Datasets} & \multirow{2}{*}{Methods}  & \multicolumn{3}{c}{R Precision$\uparrow$} & \multirow{2}{*}{MM Dist$\downarrow$} & \multirow{2}{*}{Bleu@1$\uparrow$} & \multirow{2}{*}{Bleu@4$\uparrow$} & \multirow{2}{*}{Rouge$\uparrow$}& \multirow{2}{*}{Cider$\uparrow$}& \multirow{2}{*}{BertScore$\uparrow$}\\

    \cline{3-5}
       ~ & ~ & Top 1 & Top 2 & Top 3 \\
    
    \midrule
    
    \multirow{6}{*}{\makecell[c]{Human\\ML3D}} & 
    \textbf{Real Desc} & 0.523 & 0.725 & 0.828 & 2.901 & - & - & - & - & - \\
    \cline{2-11}
    ~ & RAEs~\cite{yamada2018paired} & 0.100 & 0.188 & 0.261 & 6.337 & 33.3 & 10.2 & 37.5 & 22.1  & 10.7 \\
    ~ & Seq2Seq(Att) & 0.436 & 0.611 & 0.706 & 3.447 & 51.8 & 17.9 & 46.4 & 58.4  & 29.1 \\
    ~ & SeqGAN~\cite{goutsu2021linguistic} & 0.332 & 0.457 & 0.532 & 4.895 & 47.8 & 13.5 & 39.2 & 50.2  & 23.4 \\
    ~ & Ours w/o MT  & \underline{0.483} & \underline{0.678} & \underline{0.783} & \underline{3.124} & \underline{59.5} & \underline{21.2} & \underline{47.8} & \underline{68.3}  & \underline{34.9} \\
    ~ & Ours & \textbf{0.516} & \textbf{0.720} & \textbf{0.823} & \textbf{2.935} & \textbf{61.7} & \textbf{22.3} & \textbf{49.2}  & \textbf{72.5} & \textbf{37.8} \\
    
    \midrule
    
    \multirow{6}{*}{\makecell[c]{KIT-\\ML}} & 
    \textbf{Real Desc} & 0.399 & 0.618 & 0.793 & 2.772 & - & - & - & - & - \\
    \cline{2-11}
    ~ & RAEs~\cite{yamada2018paired} & 0.034 & 0.063 & 0.106 & 9.364 & 30.6 & 0.10 & 25.7 & 8.00  & 0.40 \\
    ~ & Seq2Seq(Att)  & \underline{0.293} & 0.450 & 0.555 & 4.455 & 34.3 & 9.30 & 36.3 & 37.3  & 5.30 \\
    ~ & SeqGAN~\cite{goutsu2021linguistic} &  0.109 & 0.345 & 0.425 & 6.283 & 3.12 & 5.20 & 32.4 & 29.5  & 2.20 \\
    ~ & Ours w/o MT  & 0.284 & \underline{0.466} & \underline{0.595} & \underline{3.979} & \underline{42.8} & \underline{14.7} & \underline{39.9} & \underline{60.1}  & \underline{18.9} \\
    ~ & Ours  & \textbf{0.359} & \textbf{0.561} & \textbf{0.668} & \textbf{3.298} & \textbf{46.7} & \textbf{18.4} & \textbf{44.2} & \textbf{79.5}  & \textbf{23.0} \\
    

    \bottomrule
    \end{tabular}
    }
    \caption{\footnotesize{Quantitative evaluation results for \textcolor{red}{motion-to-text} translation on HumanML3D and KIT-ML test sets. For each metric, the best score is highlighted in \textbf{bold}, with the second best highlighted using \underline{underscore}.}}
    \label{tab:quant_motion_to_text}
        \aftertab

\end{table*}

\beforesubsubsection
\subsubsection{Quantitative Analysis.}
\Cref{tab:quant_motion_to_text} presents the quantitative evaluation results of motion to language mapping on HumanMl3D and KIT-ML test sets. We also provide the R precision and multimodal distance of \textbf{real} descriptions for reference.

The high R precision of real descriptions also evidences the effectiveness of learned motion \& text feature extractors and R precision metric. Overall,our method clearly outperforms all baseline methods over a large margin on all datasets and metrics. RAEs~\cite{wang2020learning} suffers from limited capability on modeling long-term dependencies between 3D motion and language, thus resulting in low R precision and linguistic evaluation scores. This is mitigated by introducing attention mechanism in Seq2Seq(Att) or adversarial learning in SeqGAN, which effectively lifts the top-1 R precision up by more than 20\% on HumanML3D and 10\% on KIT-ML test sets. By utilizing motion token in our framework (ours), we can observe a obvious jump on both linguistic quality (i.e., Bleu, BertScore) and motion-retrieval precision (i.e., R precision) of generated language descriptions, which is surprisingly approaching the scores of real descriptions. 


\beforesubsubsection
\subsubsection{User Study.} Beside the aforementioned objective evaluations, a crowd-sourced
\begin{wrapfigure}{r}{0.5\textwidth}
          \vspace{-2em}
  \begin{center}
    \includegraphics[width=0.5\textwidth]{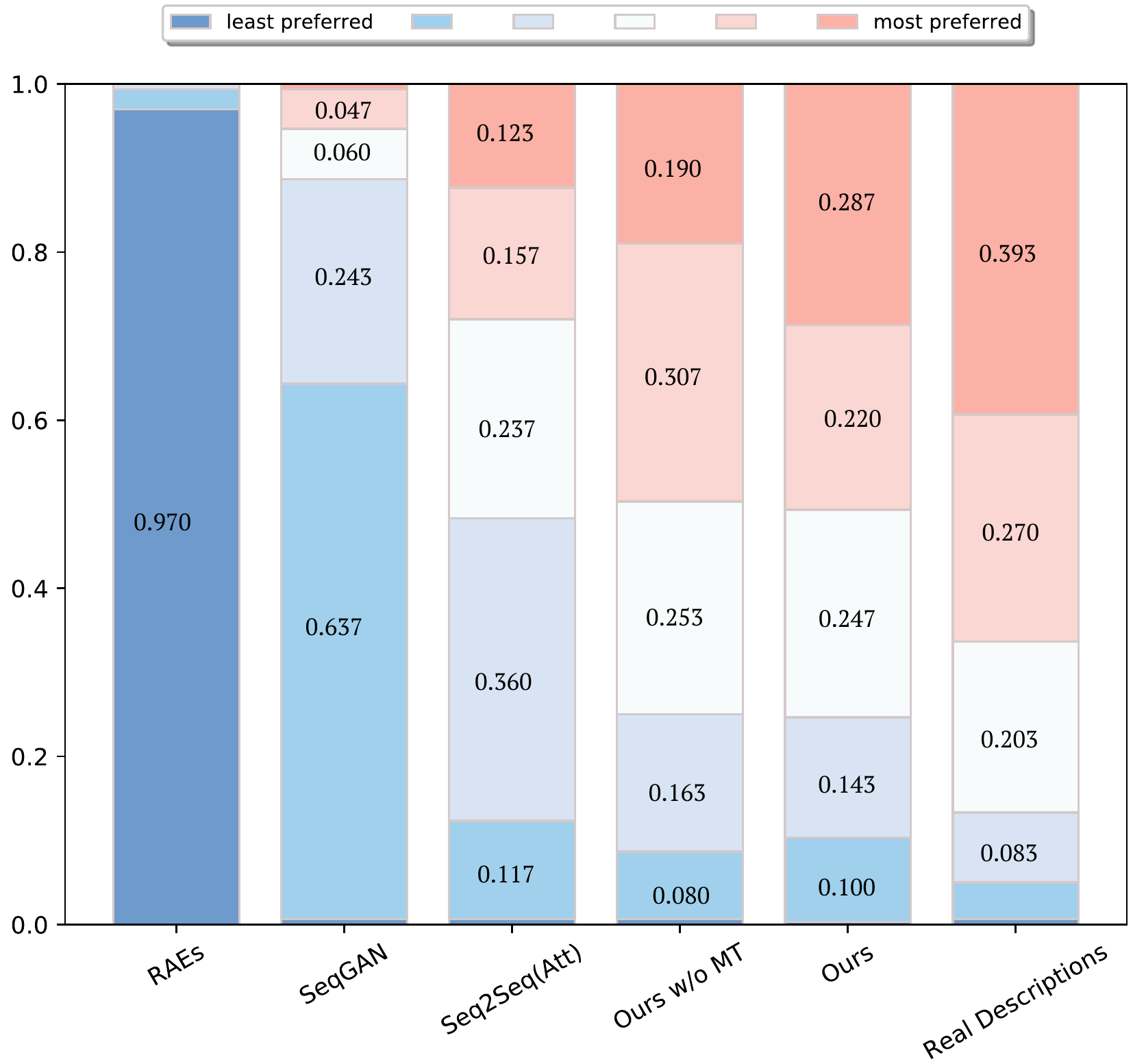}
  \end{center}
  \beforefigcaption
  \label{fig:rank_hist}
  \caption{Statistics of human preference amongst the generated descriptions for given human motions. For each method, a color bar (from blue to red) indicated the the percentage of its preference level (from least to most preferred).}
  \afterfigcaption
\end{wrapfigure}
 subjective assessment is also conducted on Amazon Mechanical Turk (AMT) involving hundreds of AMT users with \textit{master} recognition. Particularly, descriptions are generated from 100 randomly selected 3D human motions using different methods. For each human motion, the corresponding generated and real descriptions are randomly reordered and shown to 3 AMT users, who are asked to rank their preference over these descriptions based on the accuracy and fluency. 

As shown in Figure.3, our method earns the most appreciation from users over all baselines. In detail, RAEs~\cite{yamada2018paired} is the least preferred method, from which 97\% descriptions are ranked at the last place; Seq2Seq(Att) and SeqGAN~\cite{goutsu2021linguistic} gain comparably more positive feedback from users; while our method without motion tokens comes to the second to the best. This objective study solidly substantiates the capability of our approach toward generating natural as well as motion-aligned language descriptions.

\begin{figure*}[t]
	\centering
	\includegraphics[width=\linewidth]{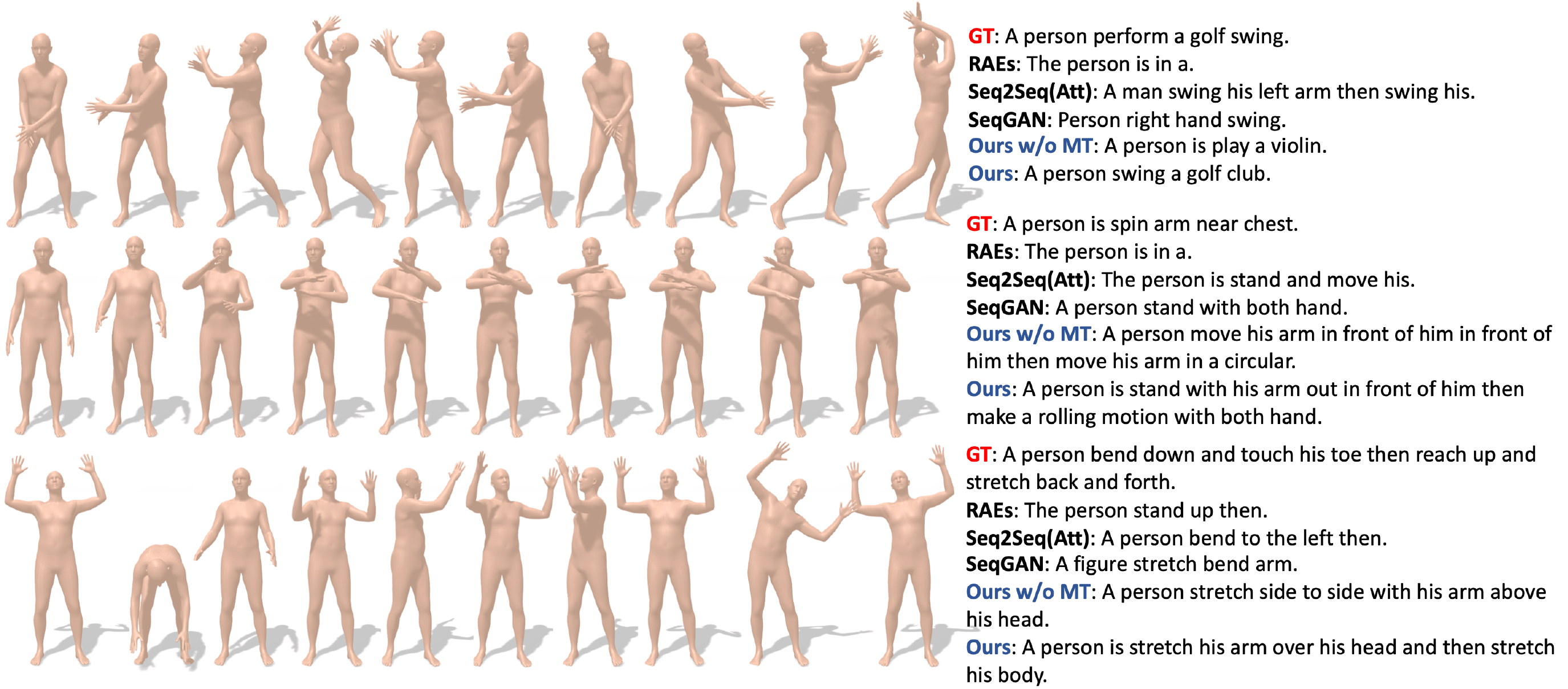}
		 \beforefigcaption
	\caption{Examples of motion-to-text translation results from different approaches. Grammatical tense and plural of words are not considered for simplifying learning process. More results are provided in supplementary files.}
	\label{fig:qualit_motion_to_text}
	 \afterfigcaption
\end{figure*}

\beforesubsubsection
\subsubsection{Qualitative Comparisons.} \cref{fig:qualit_motion_to_text} qualitatively compares the generated descriptions from different methods grounded on the same 3D human motions. RAEs~\cite{plappert2018learning} consistently produces descriptions with simple patterns like "is in a" , resulting in meaningless linguistic combinations; descriptions from Seq2Seq(Att) and SeqGAN are relatively more complex which however are usually incomplete and lack of details. Our approach without motion tokens starts to generate long and complex descriptions. Nonetheless, these descriptions sometimes fail to capture the characteristics of the input 3D motions (e.g, "play a violin"). In contrast, our approach is able to provide fluent and descriptive sentences that accurately depict various aspects of 3D motions, such as body part ("both hand"), action category ("swing", "stretch"), spatial relations ("over head").

\beforesubsection
\subsection{Evaluation of Text-to-motion Generation}
\aftersubsection
\label{exp:text-to-motion}

\begin{table*}[t]
    \centering
    \scalebox{0.78}{

    \begin{tabular}{l l c c c c c c c}
    \toprule
    \multirow{2}{*}{Datasets} & \multirow{2}{*}{Methods}  & \multicolumn{3}{c}{R Precision$\uparrow$} & \multirow{2}{*}{FID$\downarrow$} & \multirow{2}{*}{MM Dist$\downarrow$} & \multirow{2}{*}{ Diversity$\rightarrow$} & \multirow{2}{*}{MModality$\uparrow$}\\

    \cline{3-5}
       ~ & ~ & Top 1 & Top 2 & Top 3 \\
    
    \midrule
                \multirow{10}{*}{\makecell[c]{Human\\ML3D}} &

        \textbf{Real motions} & \et{0.511}{.003} & \et{0.703}{.003} & \et{0.797}{.002} & \et{0.002}{.000} & \et{2.974}{.008} & \et{9.503}{.065} & -  \\
    \cline{2-9}
        ~ & Seq2Seq\cite{lin2018generating} & \et{0.180}{.002} & \et{0.300}{.002} & \et{0.396}{.002} & \et{11.75}{.035} & \et{5.529}{.007} & \et{6.223}{.061}  & -  \\

        ~ & Language2Pose\cite{ahuja2019language2pose} & \et{0.246}{.002} & \et{0.387}{.002} & \et{0.486}{.002} & \et{11.02}{.046} & \et{5.296}{.008} & \et{7.676}{.058} & -  \\
        
        ~ & Text2Gesture\cite{bhattacharya2021text2gestures} & \et{0.165}{.001} & \et{0.267}{.002} & \et{0.345}{.002} & \et{5.012}{.030} & \et{6.030}{.008} & \et{6.409}{.071} & -  \\
        
        ~ & Hier\cite{ghosh2021synthesis} & \et{0.301}{.002} & \et{0.425}{.002} & \et{0.552}{.004} & \et{6.532}{.024} & \et{5.012}{.018} & \et{8.332}{.042} & -  \\

        ~ & MoCoGAN\cite{tulyakov2018mocogan} & \et{0.037}{.000} & \et{0.072}{.001} & \et{0.106}{.001} & \et{94.41}{.021} & \et{9.643}{.006} & \et{0.462}{.008} & \et{0.019}{.000}  \\

        ~ & Dance2Music\cite{lee2019dancing} & \et{0.033}{.000} & \et{0.065}{.001} & \et{0.097}{.001} & \et{66.98}{.016} & \et{8.116}{.006} & \et{0.725}{.011} & \et{0.043}{.001}  \\
    \cline{2-9}
            ~ & Ours baseline(T) & \ets{0.351}{.003} & \et{0.521}{.003} & \et{0.627}{.003} & \ets{1.669}{.025} & \et{4.046}{.018} & \etb{9.632}{.072} & \etb{4.352}{.149}  \\
        ~ & Ours baseline & \ets{0.351}{.002} & \ets{0.526}{.002} & \ets{0.635}{.002} & \et{1.739}{.022} & \ets{3.965}{.010} & \ets{8.651}{.083} & \ets{3.139}{.083}  \\
        ~ & Ours & \etb{0.424}{.003} & \etb{0.618}{.003} & \etb{0.729}{.002} & \etb{1.501}{.017} & \etb{3.467}{.011} & \et{8.589}{.076} & \et{2.424}{.093}  \\
    
    \midrule
            \multirow{10}{*}{\makecell[c]{KIT-\\ML}} &
 \textbf{Real motions} & \et{0.424}{.005} & \et{0.649}{.006} & \et{0.779}{.006} & \et{0.031}{.004} & \et{2.788}{.012} & \et{11.08}{.097} & -  \\
    \cline{2-9}
        ~ & Seq2Seq\cite{lin2018generating} & \et{0.103}{.003} & \et{0.178}{.005} & \et{0.241}{.006} & \et{24.86}{.348} & \et{7.960}{.031} & \et{6.744}{.106}  & -  \\

        ~ & Language2Pose\cite{ahuja2019language2pose} & \et{0.221}{.005} & \et{0.373}{.004} & \et{0.483}{.005} & \et{6.545}{.072} & \et{5.147}{.030} & \et{9.073}{.100} & -  \\

        ~ & Text2Gesture\cite{bhattacharya2021text2gestures} & \et{0.156}{.004} & \et{0.255}{.004} & \et{0.338}{.005} & \et{12.12}{.183} & \et{6.964}{.029} & \et{9.334}{.079} & -  \\
        
        ~ & Hier\cite{ghosh2021synthesis} & \et{0.255}{.006} & \ets{0.432}{.007} & \et{0.531}{.007} & \et{5.203}{.107} & \et{4.986}{.027} & \et{9.563}{.072} & -  \\

        ~ & MoCoGAN\cite{tulyakov2018mocogan} & \et{0.022}{.002} & \et{0.042}{.003} & \et{0.063}{.003} & \et{82.69}{.242} & \et{10.47}{.012} & \et{3.091}{.043} & \et{0.250}{.009}  \\

        ~ & Dance2Music\cite{lee2019dancing} & \et{0.031}{.002} & \et{0.058}{.002} & \et{0.086}{.003} & \et{115.4}{.240} & \et{10.40}{.016} & \et{0.241}{.004} & \et{0.062}{.002}  \\
    \cline {2-9}
            ~ & Ours baseline(T) & \ets{0.260}{.005} & \et{0.426}{.007} & \ets{0.538}{.008} & \ets{4.628}{.126} & \et{4.835}{.076} & \ets{12.16}{.120} & \ets{4.436}{.106}  \\
        ~ & Ours baseline & \et{0.251}{.007} & \et{0.418}{.008} & \et{0.535}{.007} & \et{4.814}{.145} & \ets{4.682}{.048} & \etb{10.13}{.117} & \etb{4.486}{.117}  \\
        ~ &Ours & \etb{0.280}{.005} & \etb{0.463}{.006} & \etb{0.587}{.005} & \etb{3.599}{.153} & \etb{4.591}{.026} & \et{9.473}{.117} & \et{3.292}{.081}  \\
    \bottomrule
    \end{tabular}
    }
    \caption{\footnotesize{Quantitative evaluation results for \textcolor{blue}{text-to-motion} mapping on HumanML3D and KIT-ML test sets. All baselines requires fixed motion lengths, and initial poses are further in demand for deterministic methods (first 4 baselines), which are all unnecessary in our approach. $\pm$ indicates 95\% confidence interval, and $\rightarrow$ means the closer to the real motion the better. For each metric, the best score is highlighted in \textbf{bold}, while the second best is hightlighted using \underline{underscore}.}}
    \label{tab:quanti_text_to_motion}
        \aftertab

\end{table*}

Mapping language to 3D human motions in a non-deterministic fashion is relatively new. Here we compare our method to four state-of-the-art methods: Seq2Seq~\cite{lin2018generating}, Language2Pose~\cite{ahuja2019language2pose}, Text2Gesture~\cite{bhattacharya2021text2gestures} and Hier~\cite{ghosh2021synthesis}. As with all existing methods, they are unfortunately deterministic methods. Therefore, two stochastic methods in other related fields are adopted here for more fair and in-depth evaluations: MoCoGAN~\cite{tulyakov2018mocogan} and Dance2Music~\cite{lee2019dancing}. MoCoGAN is widely used for conditioned video sequence synthesis, and Dance2Music learns to map sequential audio signals to 2D human dance motions. Proper changes are made to these methods for language-grounded 3D human motion generation. Ours baseline and ours baseline(T) ablates inverse alignment module during training Text2motion and map texts to motions using GRU and Transformer respectively. We repeat each experiment 20 times and report the mean value with 95\% statistical confidence interval.

\beforesubsubsection
\subsubsection{Quantitative Analysis.} \Cref{tab:quanti_text_to_motion} shows the quantitative evaluation results of language grounded 3D human motion generation. We can observe that the motions from non-determinstic baselines, MoCoGAN~\cite{tulyakov2018mocogan} and Dance2Music~\cite{lee2019dancing}, suffers from severely low quality and diversity, as reflected by their low R precision and mutimodality score. Deterministic baselines such as Seq2Seq~\cite{lin2018generating} and Text2Gesture~\cite{bhattacharya2021text2gestures} autoregressively regress human poses from textual input via vanilla sequence-to-sequence RNN and transformer respectively. However, such straightforward approaches find difficulty in maintaining textual semantics during generating human dynamics, which results in low motion-based text retrieval precision and high multimodal distance. Language2Pose~\cite{ahuja2019language2pose} and Hier~\cite{ghosh2021synthesis} propose to learn a co-embedding space between language and human motions, while Hier~\cite{ghosh2021synthesis} go one step forward by incorporating the hierarchical topology of human skeleton. These have effectively boosted the performance on both datasets. Nevertheless, there still remain a significant gap between the synthetic results and real motions. Our framework of incorporating motion token and NMT model (\textit{ours, ours baseline/baseline(T)}) in general achieve better performance, while the inverse alignment strategy greatly benefits this framework (\textit{ours}) with the top-1 and top-3 precision increased by nearly 7\% and 10\% on HumanML3D dataset.

\begin{figure*}[t]
	\centering
	\includegraphics[width=\linewidth]{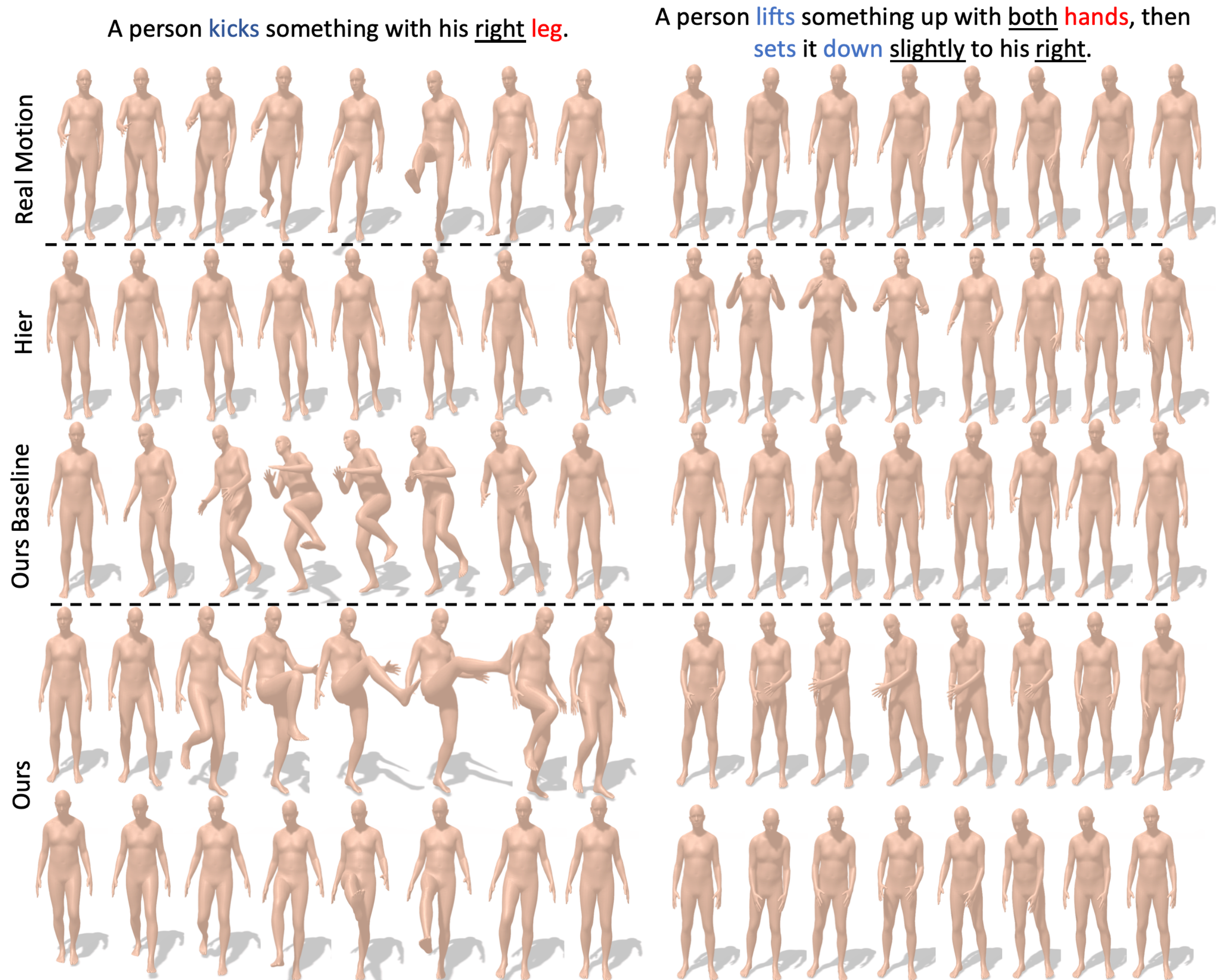}
		 \beforefigcaption
	\caption{Visual comparisons of generated motions from the same language descriptions. For each description, we show its corresponding real motion, one motion from Hier~\cite{ghosh2021synthesis} (since it's deterministic) and ours method without inverse alignment, as well as two motions from our method. Key frames of variable-length motion clips are shown. Refer to supplementary files for complete motions and more results.} 
	\label{fig:qualit_text_to_motion}
	 \afterfigcaption
\end{figure*}

\beforesubsubsection
\subsubsection{Visual Comparisons.} In \cref{fig:qualit_text_to_motion}, we visually compares the generated motions from our method (ours), our method not using inverse alignment (ours baseline), and the best performing state of the art,  Hier~\cite{ghosh2021synthesis}. The corresponding real motions are also provided for reference. Hier~\cite{ghosh2021synthesis} could somewhat capture partial concept (e.g., "kick") in descriptions, while the produced motions are unfaithfully in low-mobility. Our method without inverse alignment is capable of generating natural and plausible human motions. It sometimes however still fail to present fine details (e.g., "right leg") from texts. On the contrary, our approach consistently produce visually appealing motions which precisely convey the language concepts in descriptions.

\begin{figure*}[t]
	\centering
	\includegraphics[width=\linewidth]{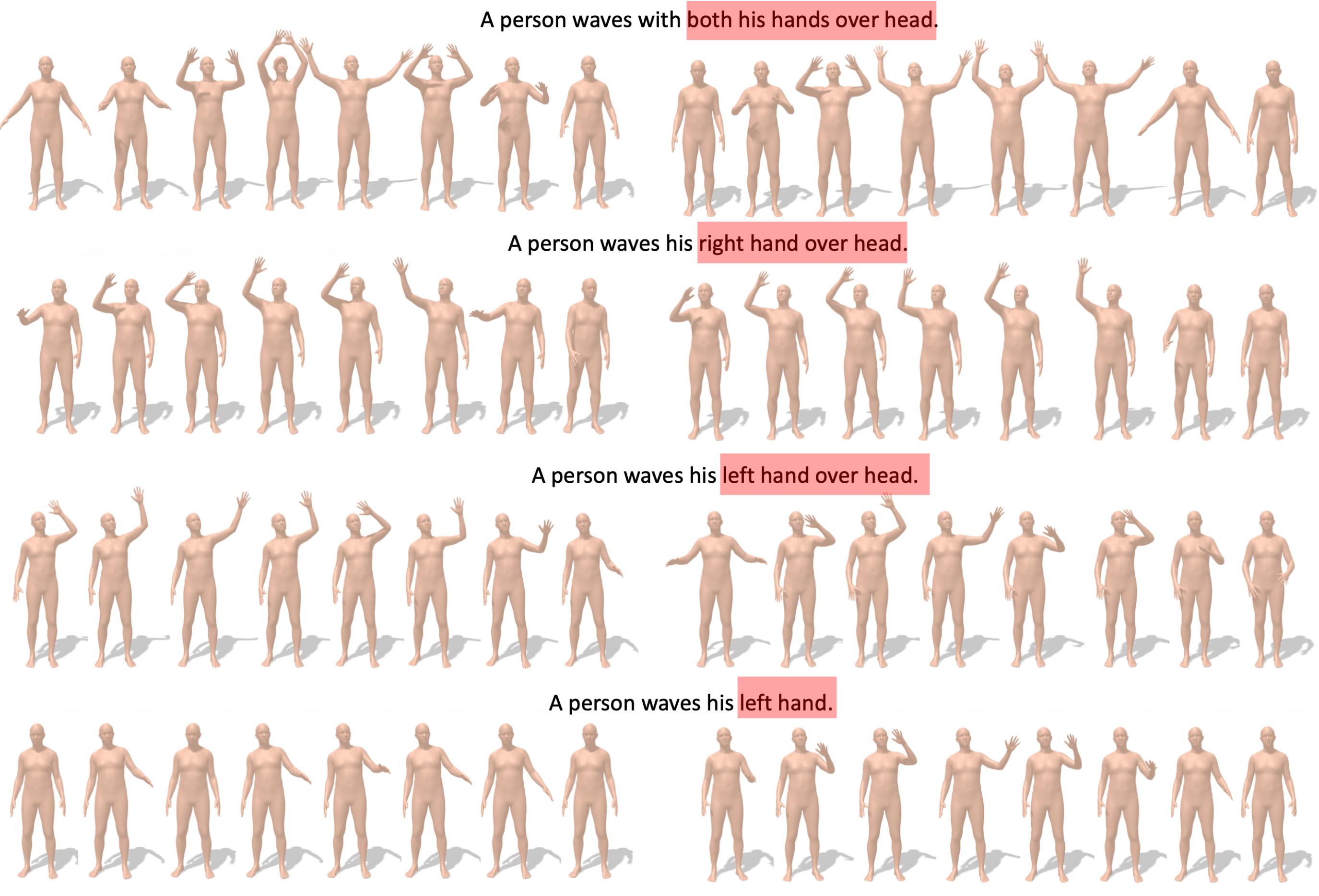}
		 \beforefigcaption
	\caption{Examples of text-to-motion mapping by modifying specific parts of text descriptions (highlighted in \textcolor{red}{red} box). For each description, we show two resultant motions.}
	\label{fig:qualit_1_motion_to_text}
	 \afterfigcaption
\end{figure*}

\beforesubsubsection
\subsubsection{Text Modifications.} We also generate 3D motions from language by modifying fixed components of the input text descriptions (\cref{fig:qualit_1_motion_to_text}). Our text2motion is able to capture the subtle semantic differences (e.g., "both/left/right hand", "over head") in text descriptions.

\beforesubsubsection
\subsubsection{Inference Time Analysis.} Time consumption of generating 300 motions from different methods on one Nvidia2080Ti: Seq2Seq (14s), Language2Pose (10s), MoCoGAN (1s), Dance2Music (1s), Text2Gesture (250s), Hier(39s), Ours (9s). Benefiting from reduced time length, our approach is able to provide the same amount of motions with even less time cost than most baselines.

\beforesubsection
\subsection{Limitations and Discussions}
\aftersubsection
Although our proposed TM2T achieves superior performance on both tasks, some limitations and potential remedies can be taken into accounts in future studies. First, the approximation in motion quantization is unfortunately not lossless, which sometimes lead to blurriness and artifacts in local body (e.g., foot sliding). Second, dealing with long and complex descriptions for text2motion is somewhat beyond our capability. This could be possibly solved by using more advanced NMT models. Third, our motion2text model is trained independently with text2motion. Learning these two mapping functions jointly and reciprocally could be another interesting topic.

\beforesection
\section{Conclusion}
\aftersection

This paper presents TM2T, a general framework that works on the bi-modal mutual mappings between 3D human motions and texts, where motion2text is further reciprocally integrated as a part of text2motion learning through inverse alignment. A new motion representation, motion token, is proposed that compress 3D motions into short sequence of discrete variables. With motion token, neural machine translation networks efficiently build mappings in-between two modalities, that is able to produces accurate descriptions as well as sharp and diverse 3D human motions. Our proposed framework is shown to produce state-of-the-art results on two motion-language dataset in both tasks.\\

\noindent\textbf{\small{ACKNOWLEDGEMENTS}}\\
This work was partly supported by the NSERC Discovery, UAHJIC, and CFI-JELF grants. I also appreciate that the university of Alberta fund me with the Alberta Graduate Excellence Scholarship.
\clearpage

%
%
\bibliographystyle{splncs04}
\bibliography{egbib}

\clearpage

\appendix
\noindent\textbf{\large APPENDIX}
\begin{abstract}
This supplementary provides more details on data pre-process, implementations details, evaluation metrics, baseline implementations, AMT user study, motion token contexts and network architecture.
\end{abstract}
\beforesection
\section{Data Preprocess}
\aftersection
\noindent\textbf{Pose Representation.} For pose representation, we extract root angular velocity, root linear velocities, root height, local joint positions, velocities, 6D rotations~\cite{zhou2019continuity} and foot contacts from raw motions as in~\cite{holden2017phase}. This results in 263 and 251 dimensional pose vectors for HumanML3D (22-joint skeleton) and KIT-ML (21-joint skeleton) dataset respectively. After all, Z-score normalization is applied to both datasets.

During training motion quantization model, to mitigate foot sliding phenomenon, the decoder $\mathrm{D}$ is asked to additionally predict foot contact information which is not provided to the encoder $\mathrm{E}$. We also scale the magnitude of root angular velocity, root linear velocities, root height and foot contacts by a value of 5 to amplify their importance. To improve the robustness of our approach, during learning motion2text and text2motion models, we randomly cutting off 0 to 4 frames at the head or tail of pose sequences, which increases the data variance while not scarifying the quality.

\beforesubsection
\section{Implementation Details}
\aftersubsection
Our framework is implemented by PyTorch. Our codebook $\mathscr{B}$ contains 1024 1024-dimentional embedding vectors. 
Encoder and decoder in motion quantization are two 1D convolutional/upsampling layers with resblocks. Weighting factor $\beta$ is set to 1. Transformers for motion2text and text2motion have 4 and 3 attention layers respectively, both with 8 attention heads with 512 hidden size. The GRU based text2motion model have encoder with hidden size of 512 while the decoder is modeled as 1-layer GRU with hidden size of 1024. This GRU model is trained with teacher force ratio of 0.4. Bi-directional GRUs with hidden size 1024 are used for motion \& text feature extractors. Adam is used for all experiments with learning rate of 0.0002. We use the codebase NLPEval~\footnote{https://github.com/Maluuba/nlg-eval} to calculate linguistic metrics (e.g., Bleu, Rouge). In text2motion, we use pre-trained 300-dimensional word embedding vectors from GloVe~\cite{pennington2014glove}. 

\beforesection
\section{Evaluation Metrics}
\aftersection

\begin{figure*}[t]
	\centering
	\includegraphics[width=\linewidth]{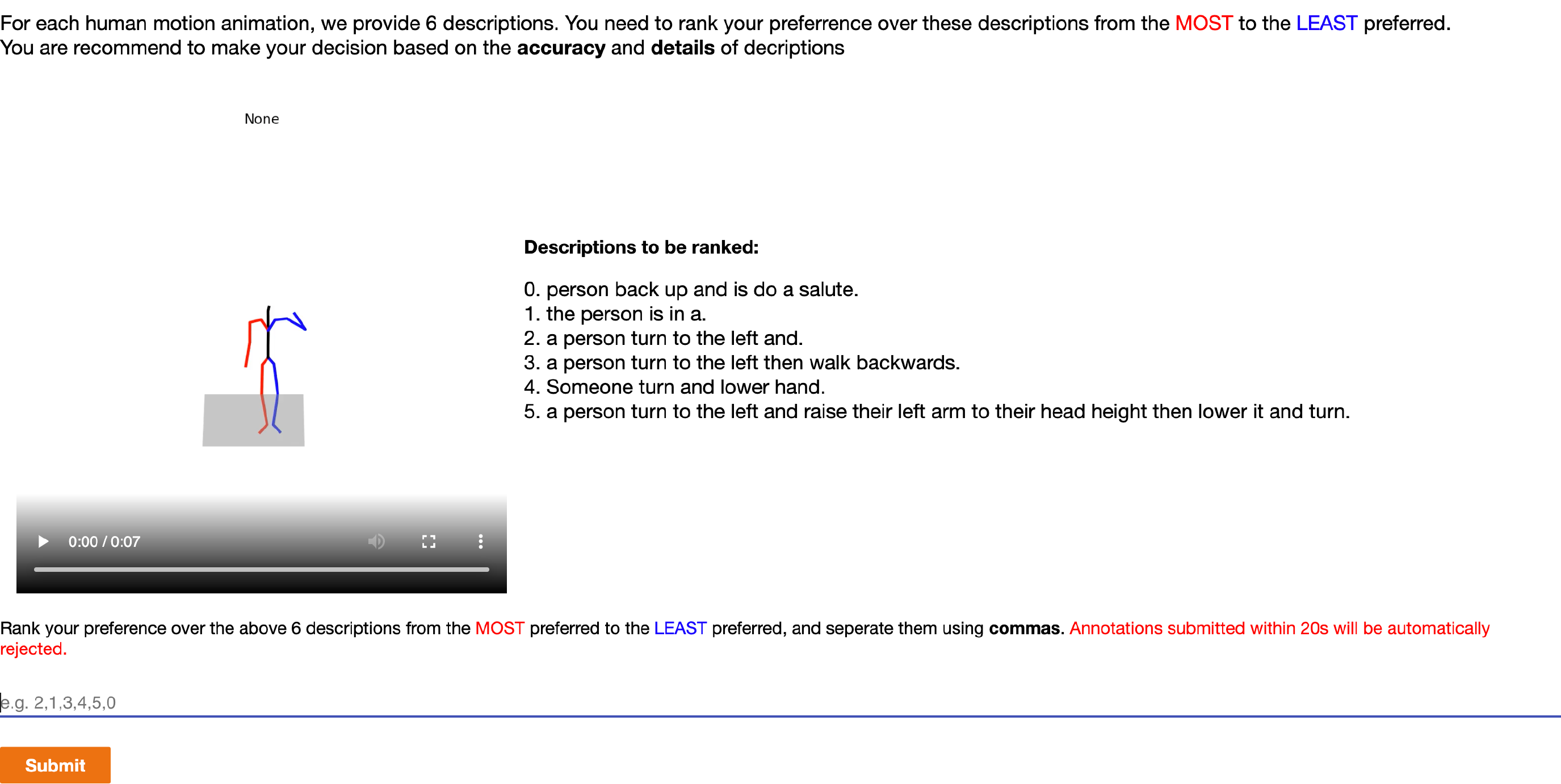}
		 \beforefigcaption
	\caption{User study interface for evaluating motion2text on Amazon Mechanical Turk.} 
	\label{fig:user_study}
	 \afterfigcaption
\end{figure*}

We first detail the process of obtaining motion and text feature extractors, and then statistical metrics for evaluating stochastic text-to-motion generation. Since metrics for motion2text translation have been well defined in existing literature~\cite{papineni2002bleu,zhang2019bertscore,lin2004rouge,vedantam2015cider}, we would like to skip the introduction for them.

\textbf{Motion and Text Feature Extractors} learn to produce geometrically closed feature vectors for matched text-motion pairs, and vice versa. Specifically, input text and motion are transformed to two semantic vectors $\mathbf{s}$ and $\mathbf{p}$ respectively using two separate bi-directional GRUs. Then, we enforce feature vectors from matched text-motion pairs to be as close as possible, while mismatched feature vectors to be separated with a margin of at least $m$. This is approached by optimizing the networks with the following contrastive loss:
\begin{align}
    \mathcal{L}_{cst} = (1-y)(\|\mathbf{s}-\mathbf{p}\|_2^2)^2 + (y)\{\max(0, m-\|\mathbf{s}-\mathbf{p}\|_2^2)\}^2,
\end{align}
where $y\in\{0, 1\}$ that $y=0$ if $\mathbf{st}$ and $\mathbf{p}$ comes from matched text-motion pairs, and vice versa. $m$ is set to 10 for both datasets. Note that test sets are untouched in this process.

The aforementioned text and motion feature extractor are then engaged in the following metrics for evaluating text2motion generation. 

\begin{itemize}
    \item \textbf{Frechet Inception Distance} (FID): Features are extracted from real motions in test set and generated motions from corresponding descriptions. Then FID is calculated between the feature distribution of generated motions vs. that of the real motions. FID is an important metric widely used to evaluate the overall quality of generated motions.
    \item \textbf{Diversity}: Diversity measures the variance of the generated motions across all descriptions. From a set of all generated motions from various descriptions, two subsets of the same size $S_d$ are randomly sampled. Their respective sets of motion feature vectors $\{\mathbf{v}_1 , ..., \mathbf{v}_{S_d}\}$ and $\{\mathbf{v}_1' , ..., \mathbf{v}_{S_d}'\}$ are extracted. The diversity of this set of motions is defined as
    
    $\mathrm{Diversity}=\frac{1}{S_d}\sum_{i=1}^{S_d}\|\mathbf{v}_i-\mathbf{v}_i'\|$
    
    $S_d=300$ is used in experiments.
    \item \textbf{MultiModality}: Different from diversity, multimodality measures how much the generated motions diversify within each text description. Given a set of motions with $C$ descriptions. For $c$-th description, we randomly sample two subsets with same size $S_m$ , and then extract two subset of feature vectors $\{\mathbf{v}_{c,1}, ... , \mathbf{v}_{c,S_m} \}$ and $\{\mathbf{v}_{c,1}', ... , \mathbf{v}_{c,S_m}' \}$. The multimodality of this motion set is formalized as
    
    $\mathrm{Multimodality}=\frac{1}{C\times S_m}\sum_{c=1}^C\sum_{i=1}^{S_m}\|\mathbf{v}_{c,i}-\mathbf{v}_{c,i}'\|$
    
    $S_m=10$ is used in experiments.
    
\end{itemize}

\beforesection
\section{Baseline Implementation}
\aftersection
For motion2text translation, unfortunately all baselines have not released their implementations yet. We re-implement SeqGAN~\cite{goutsu2021linguistic}, RAEs~\cite{yamada2018paired} and Seq2Seq(Att)~\cite{plappert2018learning} according to the descriptions in their published papers.

For text2motion generation, we re-implement Seq2Seq~\cite{lin2018generating} following its description in paper. In the official implementation of Hier~\cite{ghosh2021synthesis}, the model is trained to generate motion with fixed length (32 frames). We extend their implementation with curriculum learning to enable the motion generation with variable lengths. Proper modifications are also made to the official implementations of Text2Gesture~\cite{bhattacharya2021text2gestures} and Language2Pose~\cite{ahuja2019language2pose}, to fit in our scenario such as kinematic structure. To adapt MoCoGAN~\cite{tulyakov2018mocogan} and Dance2Music~\cite{huang2020dance} in our application, we re-use their source code and replace the categorical condition in MoCoGAN and audio signals in Dance2Music with our text features. Due to the specific architecture of their discriminator design, they are only able to generate motions with fixed lengths.
\beforesection
\section{User Study}
\aftersection

\begin{figure*}[t]
	\centering
	\includegraphics[width=\linewidth]{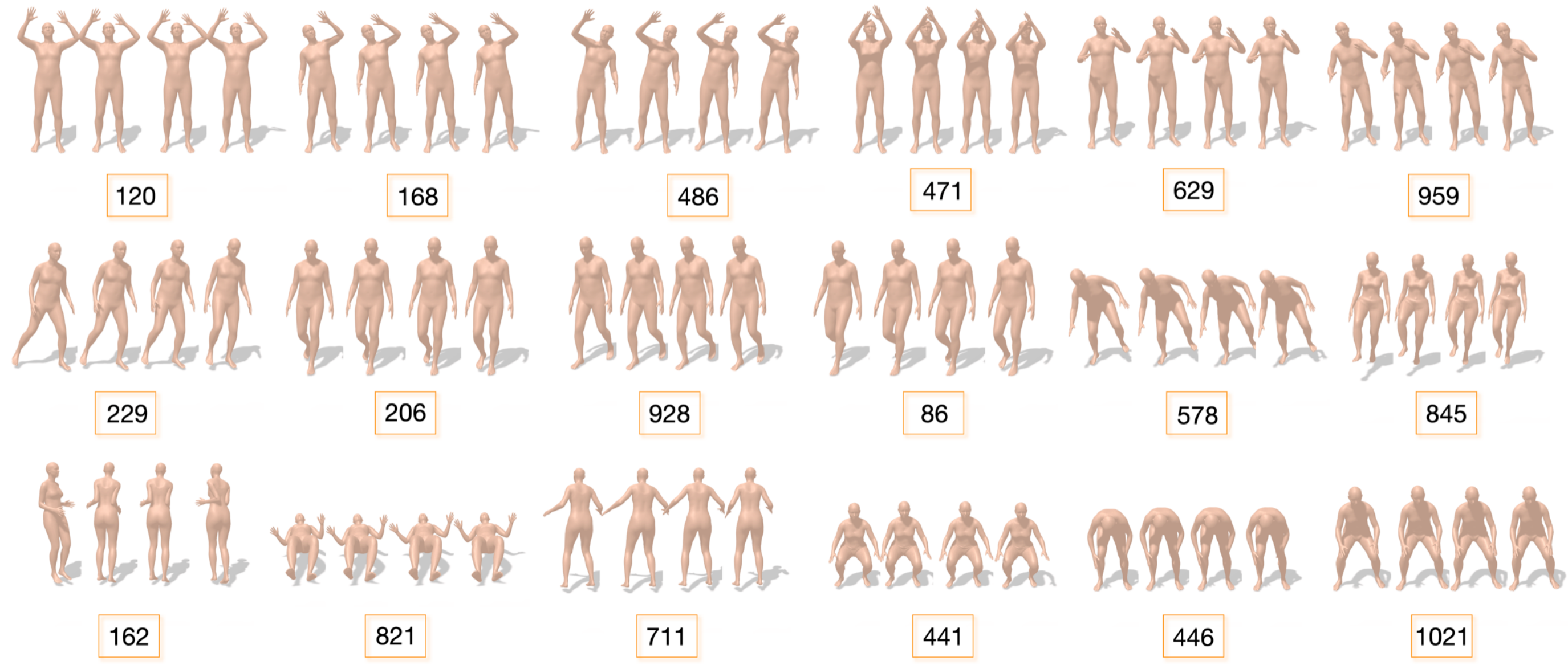}
		 \beforefigcaption
	\caption{Exemplar motion tokens and their associated local spatial-temporal contexts, visualized in 4-frame motion segments.} 
	\label{fig:token_visualization}
	 \afterfigcaption
\end{figure*}

~\cref{fig:user_study} shows the interface of our user survey for evaluating motion2text translation on Amazon Mechanical Turk. For each human motion animation, 6 generated descriptions from different source are randomly reordered. AMT users are asked to rank their preference over these 6 descriptions based on the judgement on accuracy and degree of details. Only users with \textit{master} recognition are considered.
\beforesection
\section{Motion Token Contexts}
\aftersection
To visualize the local context associated with each motion token, we decode individual tokens using the quantization decoder $\mathrm{D}$, that produces short 4-frame motion segments for each token. In ~\cref{fig:token_visualization}, we present a gallery of learned motion tokens, as well as the motion segments reflecting their contexts. Note given a tuple of motion tokens, the quantization decoder $\mathrm{D}$ learns to naturally mingle their local context with seamless transitions, rather than simply concatenating their motion segments. 

\beforesection
\section{Network Architecture}
\aftersection
~\cref{tab:architecture} elaborates the networks we are using for HumanML3D dataset. The dimension of input vectors may vary accordingly while applying to KIT-ML dataset.

\begin{table*}[ht]
  \centering
  \scalebox{0.95}{
      \begin{tabular}{l l}
        \toprule
                    Components & Architecture \\
        \midrule
        \multirow{12}{*}{\makecell[c]{Quantization \\ Encoder \\ ($\mathrm{E}$)}} & \multicolumn{1}{l}{Conv1d(259, 1024, kernel\_size=(4,), stride=(2,), padding=(1,))}\\
        & LeakyReLU(negative\_slope=0.2, inplace=True)\\
        & \makecell[l]{(ResBlock): Sequential(\\
         \,\,\, (0): Conv1d(1024, 1024, kernel\_size=(3,), stride=(1,), padding=(1,))\\
         \,\,\, (1): LeakyReLU(negative\_slope=0.2, inplace=True)\\
         \,\,\, (2): Conv1d(1024, 1024, kernel\_size=(3,), stride=(1,), padding=(1,)))}\\
        
        & Conv1d(1024, 1024, kernel\_size=(4,), stride=(2,), padding=(1,))\\
        & LeakyReLU(negative\_slope=0.2, inplace=True) \\
        & \makecell[l]{(ResBlock): Sequential(\\
         \,\,\, (0): Conv1d(1024, 1024, kernel\_size=(3,), stride=(1,), padding=(1,))\\
         \,\,\, (1): LeakyReLU(negative\_slope=0.2, inplace=True)\\
         \,\,\, (2): Conv1d(1024, 1024, kernel\_size=(3,), stride=(1,), padding=(1,)))}\\
        \midrule
        \multirow{15}{*}{\makecell[c]{Quantization \\ Decoder \\ ($\mathrm{D}$)}} & \multicolumn{1}{l}{
        \makecell[l]{(ResBlock): Sequential(\\
         \,\,\, (0): Conv1d(1024, 1024, kernel\_size=(3,), stride=(1,), padding=(1,))\\
         \,\,\, (1): LeakyReLU(negative\_slope=0.2, inplace=True)\\
         \,\,\, (2): Conv1d(1024, 1024, kernel\_size=(3,), stride=(1,), padding=(1,)))}}\\
        & \makecell[l]{(ResBlock): Sequential(\\
         \,\,\, (0): Conv1d(1024, 1024, kernel\_size=(3,), stride=(1,), padding=(1,))\\
         \,\,\, (1): LeakyReLU(negative\_slope=0.2, inplace=True)\\
         \,\,\, (2): Conv1d(1024, 1024, kernel\_size=(3,), stride=(1,), padding=(1,)))} \\
        & Upsample(scale\_factor=2.0, mode=nearest)\\
        & Conv1d(1024, 1024, kernel\_size=(3,), stride=(1,), padding=(1,))\\
        & LeakyReLU(negative\_slope=0.2, inplace=True) \\
        & Upsample(scale\_factor=2.0, mode=nearest) \\
        &  Conv1d(1024, 263, kernel\_size=(3,), stride=(1,), padding=(1,)) \\
        &  LeakyReLU(negative\_slope=0.2, inplace=True) \\
        &   Conv1d(263, 263, kernel\_size=(3,), stride=(1,), padding=(1,)) \\

        \midrule
        Codebook &  Embedding(1024, 1024)\\
        \midrule
        \multirow{2}{*}{\makecell[c]{Text (GRU) \\ Encoder}} & \multicolumn{1}{l}{ (input\_emb):Linear(in\_features=300, out\_features=512, bias=True)} \\
        & (gru): GRU(512, 512, batch\_first=True, bidirectional=True)\\
        \midrule
        \multirow{16}{*}{\makecell[c]{Motion (GRU) \\ Decoder}} & \multicolumn{1}{l}{(input\_emb): Embedding(1027, 1024)} \\
        & (z2init): Linear(in\_features=1024, out\_features=1024, bias=True) \\
        & (gru): ModuleList( (0): GRUCell(1024, 1024)) \\
        & \makecell[l]{(att\_layer): AttLayer(\\
      \,\,\,\,\,(W\_q): Linear(in\_features=1024, out\_features=1024, bias=True)\\
      \,\,\,\,\,(W\_k): Linear(in\_features=1024, out\_features=1024, bias=False)\\
      \,\,\,\,\,(W\_v): Linear(in\_features=1024, out\_features=1024, bias=True)\\
      \,\,\,\,\,(softmax): Softmax(dim=1))}\\
      & \makecell[l]{(att\_linear): Sequential( \\
      \,\,\,\,\,(0): Linear(in\_features=2048, out\_features=1024, bias=True)\\
      \,\,\,\,\,(1): LayerNorm((1024,), eps=1e-05, elementwise\_affine=True)\\
      \,\,\,\,\,(2): LeakyReLU(negative\_slope=0.2, inplace=True))}\\
        & (gru): ModuleList((0): GRUCell(1024, 1024))\\
        & (positional\_encoder): PositionalEncoding() \\
        & (mu\_net): Linear(in\_features=1024, out\_features=128, bias=True) \\
        & (trg\_word\_prj): Linear(in\_features=1024, out\_features=1027, bias=False) \\
        \bottomrule
  \end{tabular}
  }
    \caption{Architecture of our networks on dataset HumanML3D.}
      \label{tab:architecture}

\end{table*}
\clearpage
\end{document}